\documentclass[letterpaper, 10 pt, conference]{ieeeconf}  % Comment this line out if you need a4paper

\IEEEoverridecommandlockouts                            

\usepackage{booktabs}
\usepackage{amsmath}
\usepackage{algorithm}
\usepackage[noend]{algpseudocode}
\usepackage{graphics}
\usepackage{epsfig}
\usepackage[shortcuts]{extdash}
\usepackage{upgreek}
\usepackage{wasysym}
\usepackage{hyperref}
\usepackage{xcolor}
\usepackage{caption}
\usepackage{subcaption}
\usepackage{rotating}
\usepackage{amssymb}
\usepackage{multirow}
\usepackage{booktabs}
\usepackage{tablefootnote}
\usepackage{tabularx, booktabs, multirow, hyperref}
\usepackage{bm}      % For bold math symbols
\usepackage{lmodern} % For improved font support
\usepackage[T1]{fontenc}

\title{\LARGE \bf Collaborative Aquatic Positioning System Utilising Multi-beam Sonar and Depth Sensors}

\author{Xueliang Cheng\,$^{1}$, Ognjen Marjanovic\,$^{1}$, Barry Lennox\,$^{1}$, Keir Groves\,$^{1}$
\thanks{$^{1}$ Manchester Centre for Robotics and AI, University of Manchester, Manchester, United Kingdom. contact {xueliang.cheng@postgrad.manchester.ac.uk}}
}

\begin{document}

\maketitle
\thispagestyle{empty}
\pagestyle{empty}
%%%%%%%%%%%%%%%%%%%%%%%%%%%%%%%%%%%%%%%%%%%%%%%%%%%%%%%%%%%%

\begin{abstract}
Accurate positioning of underwater robots in confined environments is crucial for inspection and mapping tasks and is also a prerequisite for autonomous operations.
Presently, there are no positioning systems available that are suited for real-world use in confined underwater environments, unconstrained by environmental lighting and water turbidity levels, and have sufficient accuracy for reliable and repeatable navigation.
This shortage presents a significant barrier to enhancing the capabilities of remotely operated vehicles (ROVs) in such scenarios.
This paper introduces an innovative positioning system for ROVs operating in confined, cluttered underwater settings, achieved through the collaboration of an omnidirectional surface vehicle and an underwater ROV. 
A mathematical formulation based on the available sensors is proposed and evaluated.
%
% Experimental results from both a simulation environment and a real-world test facility, form a proof of principle of the system and demonstrate that is deployable in real word scenarios.
Experimental results from both a high-fidelity simulation environment and a mock-up of an industrial tank provide a proof of principle for the system and demonstrate its practical deployability in real-world scenarios.
Unlike many previous approaches, the system does not rely on fixed infrastructure or tracking of features in the environment and can cover large enclosed areas without additional equipment.

\end{abstract}
%%%%%%%%%%%%%%%%%%%%%%%%%%%%%%%%%%%%%%%%%%%%%%%%%%%%%%%%%%%%%

\section{Introduction}\label{sec:intro}
\subsection{Motivation} \label{sec: motivation}
Over the last decade, the domain of underwater robotics has experienced considerable expansion. Presently, the use of ROVs has become safe and routine, not only within the offshore industry, but also in confined aquatic environments. Numerous examples of their successful deployment can be found in the literature, such as pipeline inspection~\cite{zhao2022offshore}, monitoring of nuclear storage facilities~\cite{griffiths2016avexis} and liquid storage tank inspections~\cite{duecker2019learning}. Safety concerns and cost-effectiveness are the primary motivations behind the transition from human divers to ROVs~\cite{ozog2016long}.
% The use of remotely operated vehicles (ROVs) to perform underwater inspection and monitoring tasks in confined aquatic environments is now commonplace. Numerous examples can be found in the literature, such as pipeline inspection~\cite{zhao2022offshore}, monitoring of nuclear storage facilities~\cite{griffiths2016avexis} and liquid storage tank inspections~\cite{duecker2019learning}. Safety concerns and cost-effectiveness are the primary motivations behind this transition from human divers to ROVs~\cite{ozog2016long}.
However, navigating underwater vehicles through confined and sometimes cluttered environments is challenging. Precision movement is crucial in these environments, as vehicles must often navigate through narrow passages and avoid obstacles~\cite{lv2014design}.

Presently most ROV missions in confined underwater environments are teleoperated~\cite{watson2020localisation}.
However, there are several benefits to performing fully autonomous missions, including cost reduction, repeatability, and increased survey frequency~\cite{verfuss2019review}.
Aside from full autonomy, lower levels of autonomy, such as position and velocity control, can facilitate smooth and accurate navigation or accurate logging of sensor positions and trajectories for accurate surveying.
To achieve autonomy at any level, accurate positioning is essential~\cite{watson2020localisation} and currently there are no positioning systems available that are practical for use in confined underwater environments (as discussed in Section~\ref{subsection:LR}).
In addition to facilitating autonomy, location-tagged sensor data can be used for long-term environmental monitoring and analysis.
In a recent study~\cite{10018645} published from the Fukushima Daiichi Nuclear Power Station in Japan, an investigation was conducted into the environmental conditions of an underground facility used for the temporary storage of contaminated water. Hundreds of sandbags filled with zeolite particles were deployed on the floor to remove radioactivity from the water, with plans for future decommissioning using a zeolite removal ROV. The positioning of ROVs contributed to a more precise survey of the underwater environment and facilitated 3D reconstruction. However, in the current stage of deployment, ROVs have not addressed the issue of underwater localisation to achieve automation and instead, they are controlled through human intervention.
% In a recent challenge statement from the UK nuclear industry~\cite{challenge}, an accuracy requirement of 50~mm was specified for revisiting the same position in a small, enclosed storage pond of 7~m x 7~m, meaning that the localisation system must be at least this accurate because the controller that holds position and velocity will necessarily be imperfect.
% While the accuracy and precision requirements of a localisation system will vary depending on the mission and environment, the authors believe that most common tasks in enclosed environments, including autonomous navigation, could be successfully performed with a localisation system that has a root-mean-squared error (RMSE) of approximately 50~mm. Where the error is defined as the euclidean distance between the estimated position and the actual position.

To be useful in practical situations, a confined space localisation system should require minimal infrastructure, meaning that it should not rely on preinstalled cameras, beacons or sensors. Fixing infrastructure in the environment is time consuming, expensive and generally requires calibration. In addition, installing infrastructure is often not practical. For example, in a nuclear spent fuel pool, where access is highly restricted for safety reasons, it is impractical to install beacons or other infrastructure around the pool.

In this work, a Collaborative Aquatic Positioning (CAP) system is proposed that aims to satisfy the requirements stated above. 
Given the prevalent conditions of insufficient illumination and high turbidity in many underwater environments, the proposed localisation system replaces the use of optical cameras for tracking and instead employs multi-beam sonar. The proposed system uses an autonomous surface vehicle (ASV) that can self localise relative to the walls. The surface vehicle tracks and remains above the underwater ROV. By combining sensor measurements from both vehicles, the pose of the subsurface vehicle can be determined. The resulting system requires no fixed infrastructure and minimal calibration. In the next section a review of related underwater localisation systems is provided.

\subsection{Contribution}
The main contributions of this paper are as follows:
\begin{enumerate}
\item A novel formulation of a collaborative underwater positioning system that is suited for use in real world ROV missions in confined environments is presented. Unlike previous systems with comparable accuracy, the proposed system does not require any fixed infrastructure, does not rely on tracking environmental features, can cover large areas, and can operate in highly turbid water with no lighting requirements.
\item Both simulation and proof of principle studies demonstrate the correctness and effectiveness of the proposed CAP-SD mathematical model, confirming that the positioning system possesses an acceptable level of accuracy for a variety of applications.
\end{enumerate}

%%%%%%%%%%%%%%%%%%%%%%%%%%%%%%%%%%%%%%%%%%%%%%%%%%%%%%%%%%%%%
\section{Related work}\label{sec:rw}
\subsection{Literature review} 
\label{subsection:LR}

\subsubsection{ACOUSTIC TRANSPONDERS AND BEACONS} This method includes long baseline (LBL), short baseline (SBL) and ultra short baseline (USBL) systems and is primarily designed for use in open oceans.
Baseline localisation is accomplished by measuring distances between transponders and beacons based on the time-of-flight of acoustic signals.
Regarding legacy acoustic positioning systems (APS), there are primarily two types: those developed as a result of academic research and those that are commercialised products. 
In academic research, APS is often integrated with other positioning technologies, such as Doppler Velocity Loggers (DVL) and inertial sensors. In~\cite{5664570}, field experiments have demonstrated that for these technologies, the positioning error exceeds 1 meter for the majority of the time. The experimental validation of a tightly coupled USBL and inertial navigation system proposed in~\cite{morgado2013tightly} also indicates a root mean square estimation error (RMSE) in position ranging between 1 and 2 meters.
In commercial APS, ~\cite{sornadyneusbl} is among the most precise and is typically integrated into medium and small-sized underwater ROVs, offering positioning accuracy within 0.1 meters for distances up to 5 meters.

However, regardless of the type, all USBL systems require infrastructure in the form of fixed transponders to be located in the environment, and require sophisticated calibration or input from GNSS, which is often unavailable in enclosed environments.

\subsubsection{SIMULTANEOUS LOCALISATION AND MAPPING}
Simultaneous localization and mapping (SLAM) is a widely researched topic in the field of robotics for ground and aerial vehicles.
However, underwater SLAM is more challenging due to the fact that commonly used sensors such as cameras and LiDARs do not perform well underwater due to higher photon absorption.
In~\cite{rahman2022svin2}, Rahman et al. validated three visual SLAM techniques underwater, OKVIS (Stereo)~\cite{leutenegger2015keyframe}, VINS-Mono~\cite{qin2018vins} and ORB-SLAM3 (Stereo-in)~\cite{campos2021orb}.
Their work demonstrated that the visual SLAM algorithms could provide localisation with reasonable accuracy: RMSE 0.1-0.3~m.
However, visual SLAM needed to constantly identify features in the image to provide accurate positioning, which is problematic in underwater environments, where visibility is lower than in air and environments are often feature sparse.

Sonar SLAM in an underwater environment is analogous to 2D LiDAR based SLAM in a terrestrial environment~\cite{ribas2006slam}. Testing in an abandoned marina, Mallios~et~al. found that their pose-based~SLAM method had a mean accuracy of 1.9~m \cite{mallios2016toward}, which, even if this could be repeated in a confined environment, is too high to be of use in many applications.

\subsubsection{FIXED INFRASTRUCTURE CAMERA TRACKING SYSTEMS}
Duecker~et~al.~\cite{9341051} placed an array of 63 fiducial markers around the perimeter of a tank and used a vehicle mounted camera combined with AprilTag tracking to estimate the pose of the camera in the tank, achieving an RMSE of 2.6~cm.
Other systems, such as the Qualisys Miqus, use fixed cameras to track markers that are attached to the ROV and can return sub centimetre accuracy~\cite{qualisysycam}.
Despite their high accuracy, camera tracking systems with fixed infrastructure are generally only useful in lab settings, as they have significant setup and access requirements and limited volume coverage. Furthermore, since such methods primarily utilize optical cameras, they become unsuitable in underwater environments with low ambient illumination levels and high turbidity.

\subsubsection{DEAD RECKONING}
Dead reckoning estimates current position on a previously known or estimated position, and then using the current direction and distance travelled to update this position.
Pure dead reckoning uses exclusively ROV mounted sensors and requires no infrastructure.
There are two key technologies that are typically used for underwater dead reckoning: Doppler Velocity Logs (DVLs)~\cite{snyder2010doppler} and inertial measurement units (IMUs)~\cite{fukuda2021performance}; these two technologies are commonly combined~\cite{9312483}.
However, due to the nature of the method, all dead reckoning systems accumulate errors~\cite{8910622} and are therefore most commonly used to aid other systems that can directly provide position relative to the environment or fixed infrastructure.
In~\cite{rogne2016mems}, an IMU based nonlinear observer for dead reckoning was proposed.
It was demonstrated the RMSE of the system was 100 meters after 10 minutes.

\subsubsection{SUMMARY}
It is apparent that although individual technologies meet some of the requirements detailed in Section~\ref{sec: motivation}, none of the technologies reviewed are sufficient to meet the combination of requirements. From a technical perspective, relying solely on a multi-beam sonar is insufficient for determining the three-dimensional coordinates of an object~\cite{sung2020underwater}~\cite{aykin2013forward}. These motivate the development of the CAP system presented in this paper, which integrates a multi-beam sonar and depth sensor (CAP-SD).

%%%%%%%%%%%%%%%%%%%%%%%%%%%%%%%%%%%%%%%%%%%%%%%%%%%%%%%%%%%%%%%%%%%

\section{CAP-SD formulation}\label{sec:formulation}
\subsection{Notations and coordinate frames}
\begin{figure*}[ht]
    \centering
    \includegraphics[width=\textwidth]{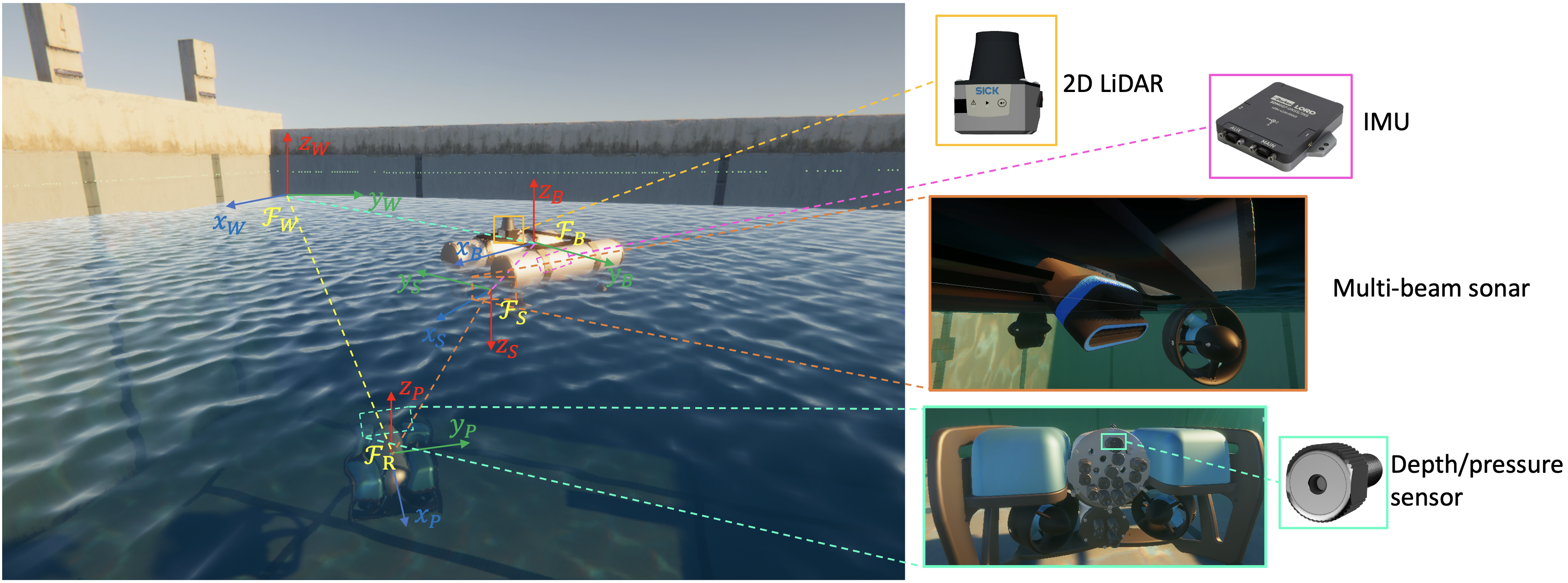}
    \caption{Overview of the proposed CAP-SD system. An omnidirectional ASV - MallARD, with integrated LiDAR, IMU, and an image sonar tracks a submersible ROV. The ROV is also fitted with a pressure sensor for depth measurement.}
    \label{fig:Structure of the system}
\end{figure*}

Figure~\ref{fig:Structure of the system} presents the coordinate system and reference frames used in this work.
The complete positioning system is composed of the following co-ordinate frames:
world frame,~$\mathcal{F}_W$,
ASV baselink frame,~$\mathcal{F}_B$,
multi-beam sonar frame,~$\mathcal{F}_S$ and
underwater ROV frame,~$\mathcal{F}_R$.
The origin of ~$\mathcal{F}_W$ has been assigned to a corner of the testing tank.
$\mathcal{F}_B$ is the geometric centre of the ASV and ~$\mathcal{F}_S$ has a fixed transform from $\mathcal{F}_B$.

The transformation from one arbitrary coordinate frame,  $\mathcal{F}_A$, to another, $\mathcal{F}_B$, can be expressed as a homogeneous transformation matrix:

\begin{equation}
\mathbf{H}^A_B=\left[\begin{array}{cc}
\mathbf{R}^A_B & \mathbf{p}^A_B \\
0 & 1
\end{array}\right]
\end{equation}

where $\mathbf{R}^A_B \in \mathrm{SO}(3)$ is a rotation matrix from frame A to frame B and $\mathbf{p}^A_B \in \mathbb{R}^3$ is a translation column vector.
% We denote the vector and scalar parts of a quaternion by $\underline{\mathbf{q}} \triangleq\left[\boldsymbol{\varrho}^T q_4\right]^T$.
$\mathbf{H} \in SE(3)$ represents the combination of rotation and translation in three dimensions, which can also be expressed as the pair $[\mathbf{R}, \mathbf{p}]$.

\subsection{Estimating the sonar pose in the world frame}
\label{sec:sonar_in_world}

The proposed CAP-SD system initially performs self-localisation of the ASV by combining 2D SLAM with IMU (Inertial Measurement Unit) data, to ascertain the pose of the ASV relative to the world-fixed frame. Details of the method for doing this have been presented in a previous paper~\cite{Cheng_2024}
by the author, with an overview provided below.

The position and orientation data \([x^W_B, y^W_B, \psi^W_B]^{\top}\) provided by SLAM is notably accurate, as verified by our prior research~\cite{groves2019mallard}. The fusion of SLAM and IMU data results in Euler angles \([\psi^W_B, \theta^W_B, \phi^W_B]^{\top}\), with pitch (\(\theta\)) and roll (\(\phi\)) derived from a tilting EKF. Consequently, the rotation matrix, \(\mathbf{R}^W_B\), is formulated using the conventional Z-Y-X Euler angles based on the fused orientation data.

To localise the ASV within the world-fixed frame, a homogeneous transformation, $\mathbf{H}^W_B = [\mathbf{R}^W_B, \mathbf{p}^W_B]$, is calculated. Utilising a tilting EKF, IMU and SLAM data are fused to estimate the USV's orientation, $\mathbf{R}^W_B$. This process incorporates tilt angles from the IMU and the yaw angle from SLAM, adopting a Z-Y-X Euler sequence for the comprehensive construction of the rotation matrix. Despite the susceptibility of Euler angles to gimbal lock, such occurrences are mitigated here by ensuring the $y$-axis rotation does not exceed 90 degrees, a condition met by surface vehicles. 

The EKF was used to accurately estimate pitch and roll angles ($\psi^W_B, \theta^W_B$) under dynamic conditions, utilising data from a three-axis gyroscope and accelerometer. Inputs to the tilting EKF include tri-axial angular rates, \(\boldsymbol{\omega} = [{\omega}_x, {\omega}_y, {\omega}_z]^{\top}\), and accelerations, \(\boldsymbol{a} = [{a}_x, {a}_y, {a}_z]^{\top}\), from the USV's gyroscope and accelerometer, respectively, with the assumption of uncorrelated Gaussian noise. The model assumes negligible non-gravitational linear accelerations during the USV's motion.

The pose of the sonar with respect to ($\mathcal{F}_W$) can be expressed by:

\begin{equation}
\label{eq:sonar pose}
\mathbf{H}_S^W =\mathbf{H}_B^W \mathbf{H}_S^B,
\end{equation}

\subsection{Identifying underwater ROV's pixels in sonar scans}
\label{sec:nn_detector}

\begin{figure}[ht]
    \centering
    \includegraphics[width=\columnwidth]{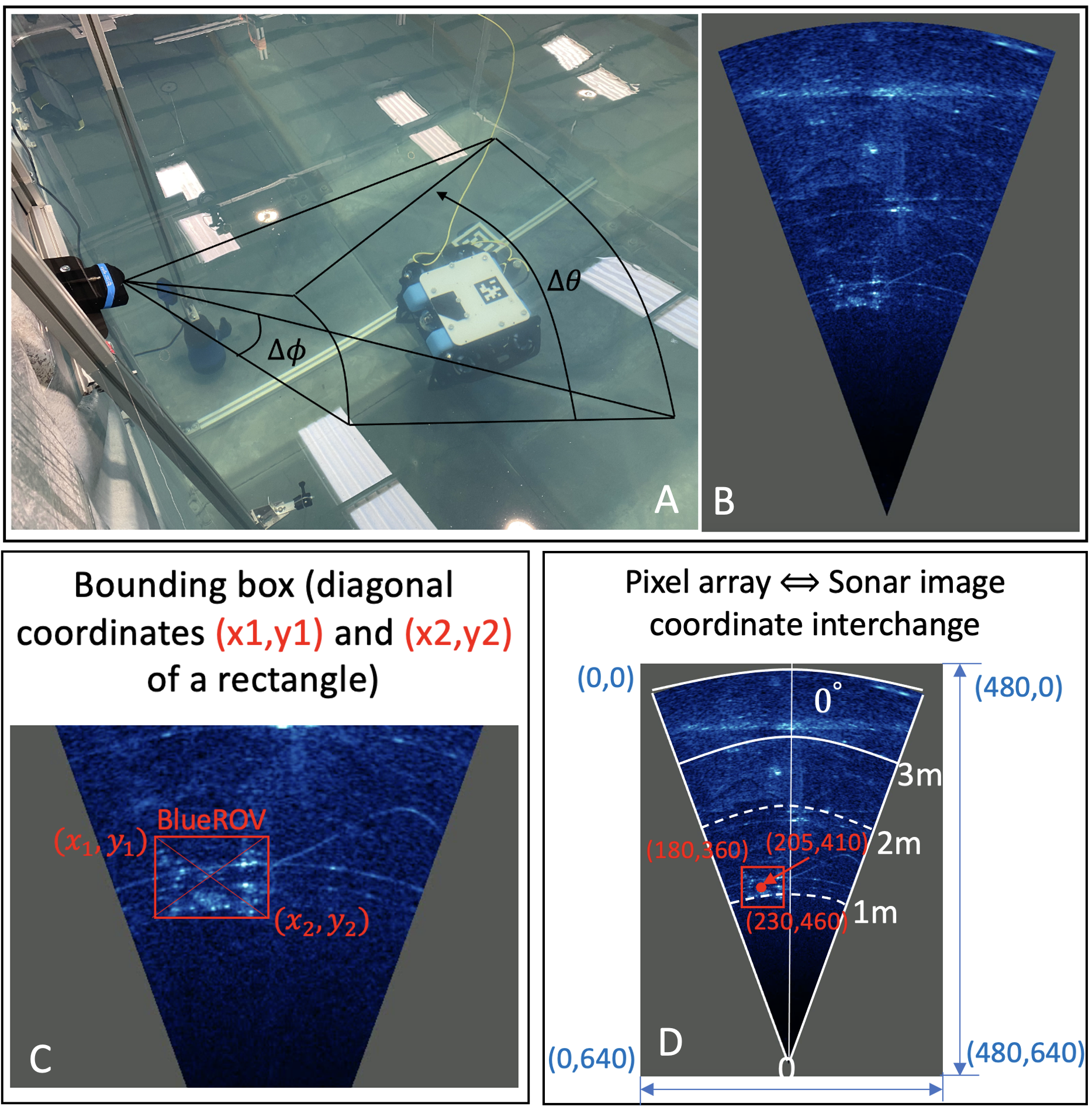}
    \caption{NN object detector (A)~Side view of a working scenario involving the commercially available BlueROV ROV, equipped with multi-beam sonar (B)~ Raw sonar image (C)~Sonar image within NN object detector (D)~Pixel array and sonar image coordinate interchange}
    \label{fig:nndetect}
\end{figure}
A high-frequency multi-beam FS sonar emits beams with a fixed vertical width defined by $\left\{\phi_{\min }, \phi_{\max }\right\}$ across various azimuth directions $\theta$. In the imaging of the multi-beam sonar, it is displayed as shown in Fig~\ref{fig:nndetect}A.

A NN detector serves to pinpoint the ROV's pixel coordinates within the sonar image. Given the predefined horizontal field-of-view (FOV), maximum range, and image resolution of an actual multi-beam sonar, determining the ROV's pixel coordinates in the sonar image consequently specifies its distance \(R\) and azimuth angle \(\theta\) in relation to the multi-beam sonar.

A pre-trained YOLOv5 model~\footnote{The detail about the customized pre-trained model can be found in:~\url{https://github.com/Xue1iang/CAP_SD_Unity}} was employed to detect the locations of any ROVs in the sonar images. This is achieved by identifying any ROVs within bounding boxes as shown in Fig~\ref{fig:nndetect}C. YOLOv5 builds on the previous YOLO version, with improvements introduced to further boost performance and flexibility to ease deployability in real-world experiments. The pixel coordinates of any ROV are obtained at this stage. Subsequently, a conversion from pixel coordinates to sonar image coordinates is performed, as depicted in Fig~\ref{fig:nndetect}D. For the sonar image frame, its coordinate parameters, such as range and vertical width, are predetermined and fixed. Therefore, pixel coordinates can be directly converted into the sonar image's $\left\{R, \theta\right\}$ coordinates.

\subsection{Estimate underwater ROV position in world frame}
\label{sec:bluerovinworld}

\begin{figure}[ht]
    \centering
    \includegraphics[width=\columnwidth]{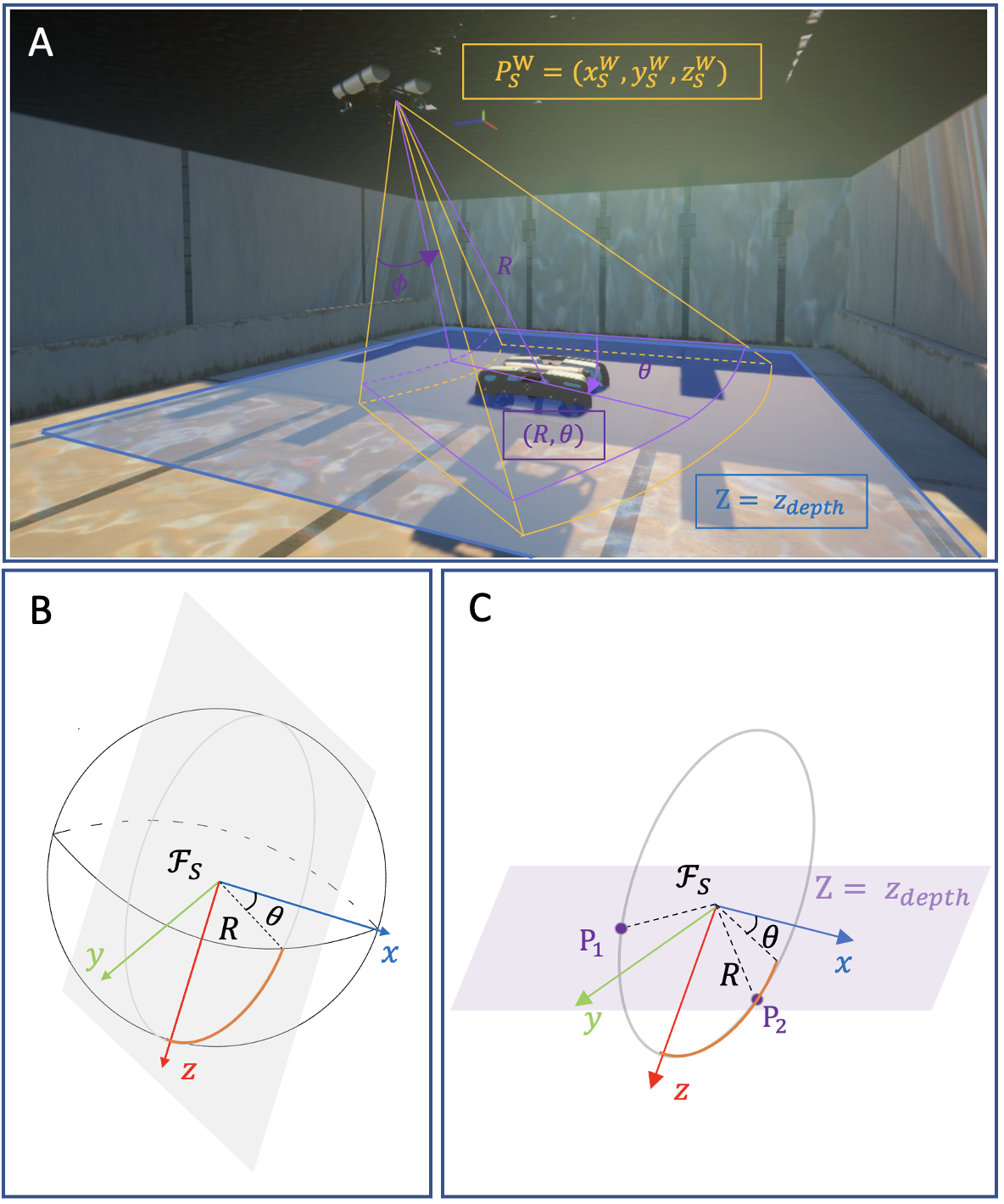}
    \caption{Abstract frames of the CAP-SD system. (A)~Underwater side-view of CAP-SD system. (B)~Using the cutting-sphere plane and sphere to define the circle where the ROV may be located. (C)~Possible locations of ROV can be reduced to two points~($P_1$ and $P_2$) by generating another plane using the depth sensor. The position of ROV ($P_2$) must be in the sonar's positive x region~(orange arc in (B) and (C)).}
    \label{fig:frames}
\end{figure}
Based on $H^W_S$ mentioned in Section~\ref{sec:sonar_in_world} and $\left\{R, \theta\right\}$ discussed in Section~\ref{sec:nn_detector}, a circle of possible positions for the ROV can be determined by solving the intersection between a sphere and a cutting-sphere plane:
\begin{equation}
\label{eq:sphe}
\left(x-x_S^W\right)^2+\left(y-y_S^W\right)^2+\left(z-z_S^W\right)^2=R^2,
\end{equation}
where $x_S^W$, $y_S^W$ and $z_S^W$ is the position of multi-beam sonar. The cutting-sphere plane containing the circle can be represented as (gray plane depicted in Figure~\ref{fig:frames}B):
\begin{equation}
A\left(x-x^W_S\right)+B\left(y-y^W_S\right)+C\left(z-z^W_S\right)=0,
\label{eq:arcplane}
\end{equation}
where $\vec{n}_{3 \times 1}=[A, B, C]^{\top}$ is the normal vector to the cutting-sphere plane, defined as:
\begin{equation}
\left[\begin{array}{c}
\vec{n}_{3 \times 1}\\
1
\end{array}\right]=(H^W_B H^B_S)^{\prime}\left[\begin{array}{c}
\sin (\theta) \\ -\cos (\theta) \\ 0 \\
1
\end{array}\right].
\end{equation}

To determine the position of the ROV in ~$\mathcal{F}_W$, it is necessary to utilise a depth sensor that can define a $z$ plane in~$\mathcal{F}_W$ :
\begin{equation}
Z=z_{depth}.
\end{equation}

% By intersecting the circle (gray circle in Figure~\ref{fig:frames}C), derived from solving Equation~\ref{eq:sphe} and~\ref{eq:arcplane}, with the plane defined by the depth sensor, two points can be obtained.
%
The plane (purple plane in Figure~\ref{fig:frames}C) defined by the depth sensor, cuts through the circle (gray circle in Figure~\ref{fig:frames}C) of possible ROV positions, resulting in two intersection points.
However, the ROV that is to be located, must appear within the FOV of the image sonar (sonar's positive x region as shown in the orange arc in Figure~\ref{fig:frames}C). 
Therefore, a single point is ultimately identified, which represents the position of the ROV $\mathbf{P}^W_R=[x^W_R, y^W_R, z^W_R]^{\top}$ with respect to ~$\mathcal{F}_W$.

\section{Results and evaluation}\label{sec:experiments}
% \subsection{Simulation architecture}
% \begin{figure}
%     \centering
%     \includegraphics[width=\columnwidth]{figures/simulation_diagram.png}
%     \caption{CAP-SD simulation communication architecture}
%     \label{fig:simulation_dia}
% \end{figure}
% The simulation stack, which incorporates a ROS backend. On the ROS~\cite{roswiki} side, a gRPC server endpoint is established, functioning as a ROS node responsible for the bidirectional distribution of data flow. On MARUS side (Unity), it obtains inputs for actuators, mission control variables, and requests for simulating communication (either optical or acoustic) within the simulated environment from ROS.
\subsection{Simulation scenario}
The CAP-SD system firstly underwent evaluation within a custom facility set up using the Unity~\cite{UnityHomepage} underwater simulation environment. Portions of the simulation, such as sensors and communication framework, utilized MARUS~\cite{lonvcar2022marus}, while the hydrodynamics components for ASVs and AUVs were implemented using the Dynamic Water Physics~\cite{DynamicWaterPhysics2020} asset in Unity.

\begin{figure}[ht]
    \centering
    \includegraphics[width=\columnwidth]{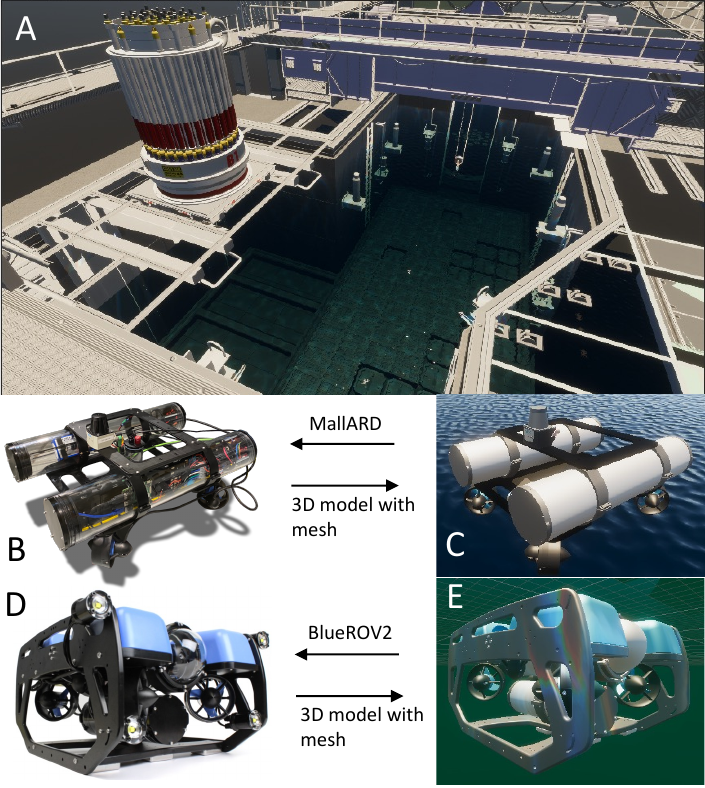}
    \caption{Simulation environment scenario and ROVs. (A)~High-level view of a simulated nuclear fuel pool. (B)~Autonomous surface robot - MallARD. (C)~MallARD's 3D model in Unity. (D)~Commercial underwater ROV - BlueROV2 (E)~BlueROV2 3D model in Unity.}
    \label{fig:experiemntal_set}
\end{figure}

Given that the application scenario of CAP-SD will be confined to indoor environments, the simulation scenario was set within a typical nuclear fuel pool (NFP). The entire area measured 28~m in length, 16~m in width, and had a depth of 8~m, as depicted in Fig~\ref{fig:experiemntal_set} A and B. 
Within the NFP, the MallARD~\cite{groves2019mallard} (shown in Fig~\ref{fig:experiemntal_set}C and D), an ASV developed by the University of Manchester and equipped with simulated multi-beam sonar, IMU, and LiDAR, was utilized alongside the commercially available BlueROV2 (shown in Fig~\ref{fig:experiemntal_set}E and F), which served as the ROV that was to be positioned. 
%
% This strategic selection was aimed at facilitating the transition of CAP-SD from simulation to real-world deployment.

\subsection{Simulation evaluation of CAP-SD}
The simulation platform was a personal computer with an Intel Core i7 12700H CPU (central processing unit) running at 3.972 GHz and with 32-GB RAM (random access memory). In the experiments, the BlueROV2 moved in the tank while maintaining a fixed pitch and roll. In the following research, MallARD was programmed to follow the BlueROV2 autonomously, with the focus of the present research is to validate and quantify the accuracy of the proposed positioning system. All sensor data was collected via ROS and was recorded on the same roscore, which allowed data to be shared in real-time and synchronised to a single clock.

To evaluate the performance of the CAP-SD system, five distinct datasets with varying motion patterns were collected. In these five datasets, the BlueROV2 manoeuvred itself along five distinct trajectories: square, lawnmower, bouncing, random, and two-floor. All datasets were recorded when MallARD was held stationary and BlueROV2 was moved underneath in the multi-beam sonar's FOV. Data from all sensors was recorded and was post processed using CAP-SD. Pose information from the Unity transformation component was recorded as the ground truth.

\subsection{Simulation results discussion}
\begin{figure}[ht]
    \centering
    \includegraphics[width=\columnwidth]{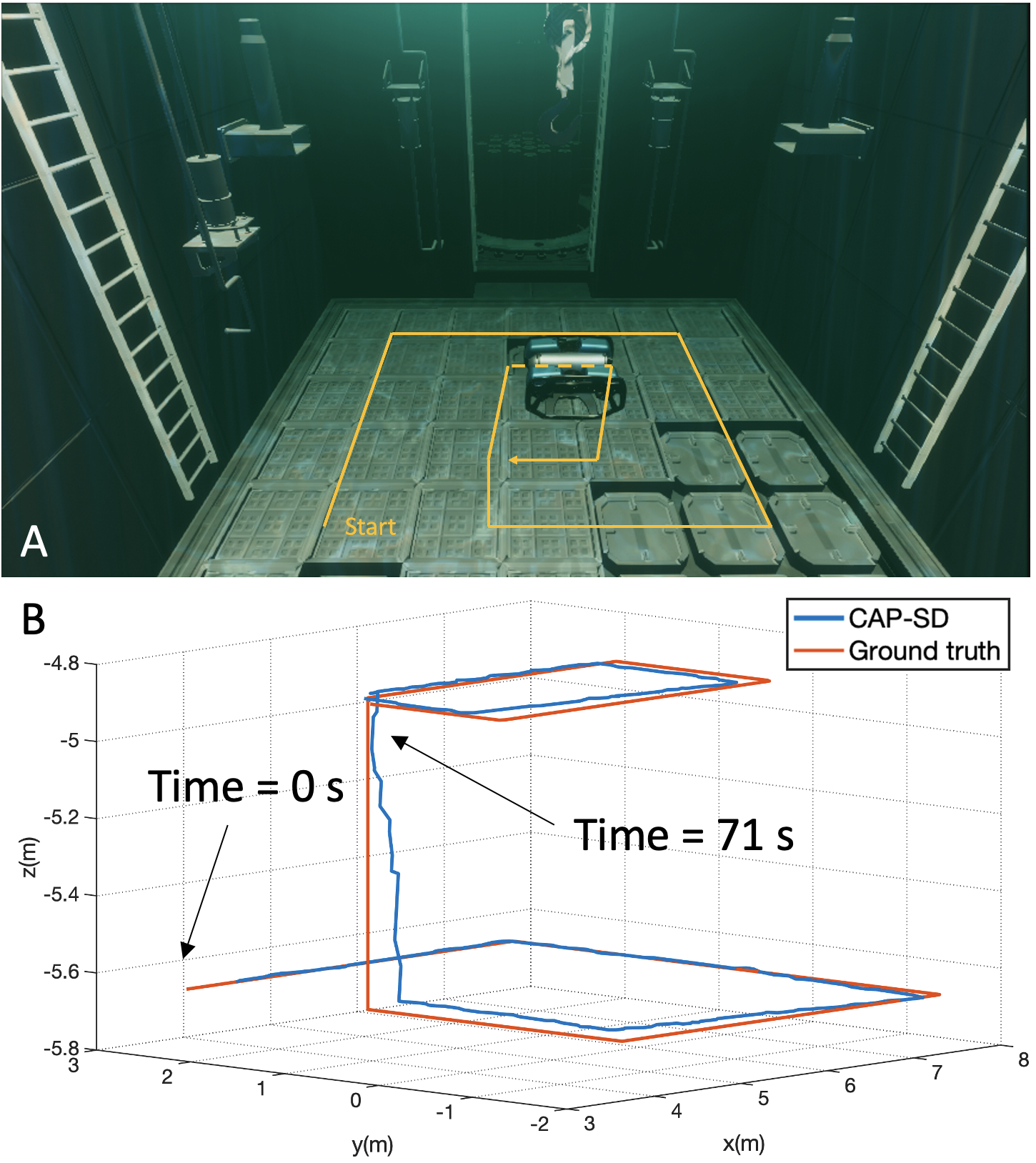}
    \caption{The 3D trajectory of BlueROV2 estimated by CAP-SD against ground truth overtime respectively, during 71 seconds of dataset 5 (Two-floor).}
    \label{fig:3d_traj_flault}
\end{figure}

\begin{figure}[ht]
    \centering
    \includegraphics[width=\columnwidth]{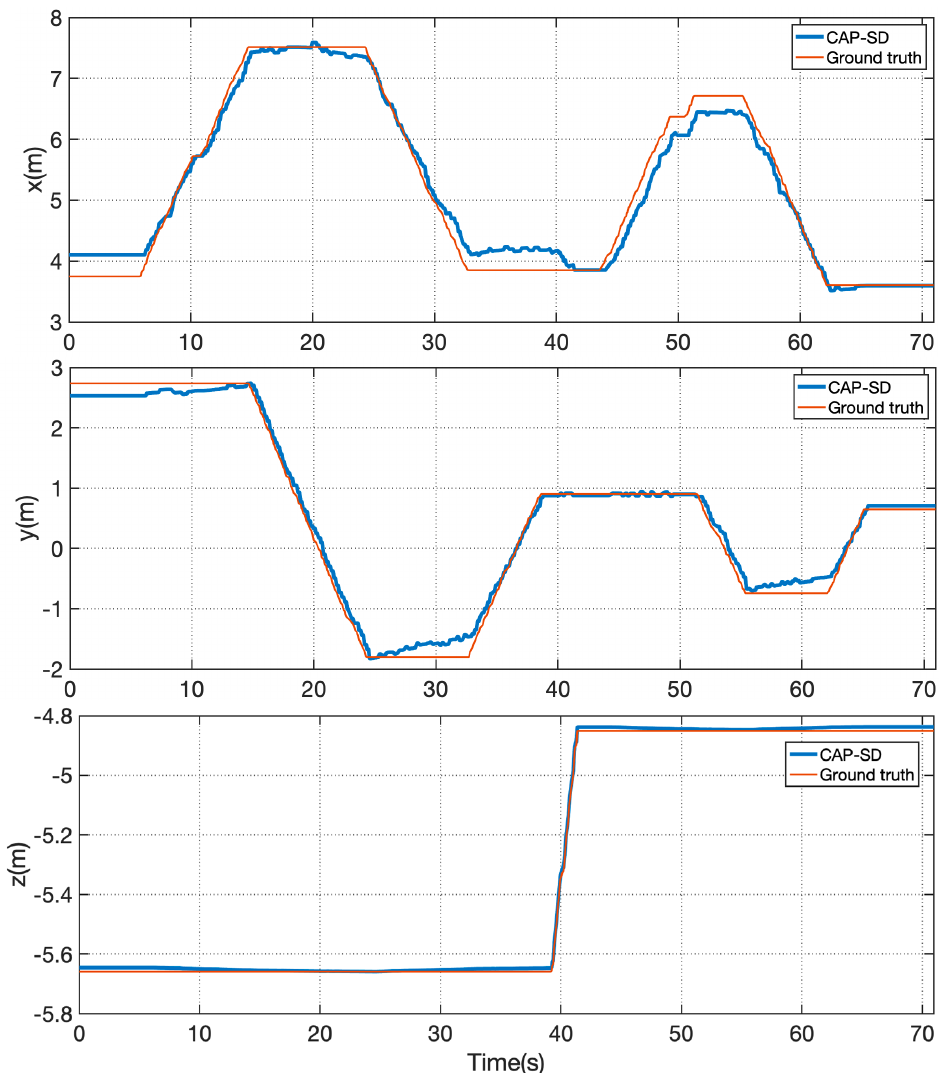}
    \caption{The positioning simulation results of the CAP-SD system from the X, Y, and Z axes}
    \label{fig:xyz_sim}
\end{figure}

Figure~\ref{fig:3d_traj_flault}B shows the 3D trajectory of the CAP-SD alongside the ground truth, for one of the motion patterns.
Overall, within this simulation environment, CAP-SD was capable of localising the underwater ROV with reasonable accuracy. Table~\ref{tab: errors_single_axis} presents the RMSE and mean euclidean distance (MED) of the proposed CAP-SD methods for all five datasets. For simplicity, the depth/pressure plugin that was used in this simulation, added noise to a known absolute depth, and hence the results do not include a comparison of the RMSE along the $z$-axis. Therefore, as detailed in Section~\ref{sec:formulation}, the accuracy of localisation largely depends on whether the YOLOv5 model can accurately determine the bounding box (the pixel coordinates of the ROV on the sonar image frame). It is noted that when generating the real results, presented in Section XXX, the depth measurement was determined directly using a differential pressure cell attached to the ROV. 
\begin{table}[ht]
\caption{RMSE and MED relative to the ground truth measurement of CAP-SD for all 5 datasets~\tablefootnote{All datasets are available in:~\url{https://github.com/Xue1iang/IROS2024_CAP-SD}}. \label{tab: errors_single_axis}}
\centering

\begin{tabularx}{\columnwidth}{Xcccc}
\toprule 
{\small{}Dataset}        & {\small{}Trajectory} & {\small{}x RMSE}  & {\small{}y RMSE} & {\small{}MED}  \\ 
\midrule
\multirow{1}{*}{1} & Square     & 134.2~mm & 129.6~mm & 160.1~mm \\  
\multirow{1}{*}{2} & Bouncing     & 164.7~mm & 144.5~mm  & 201.7~mm \\ 
\multirow{1}{*}{3} & Lawnmower     & 162.6~mm  &147.3~mm & 206.5~mm  \\ 
\multirow{1}{*}{4} & Random        & 143.9~mm  & 134.7~mm & 180.2~mm  \\ 
\multirow{1}{*}{5} & Two-floor     & 218.3~mm  & 140.3~mm & 228.3~mm  \\ 
\bottomrule
\end{tabularx}
\end{table}

\begin{figure}[ht]
    \centering
    \includegraphics[width=\columnwidth]{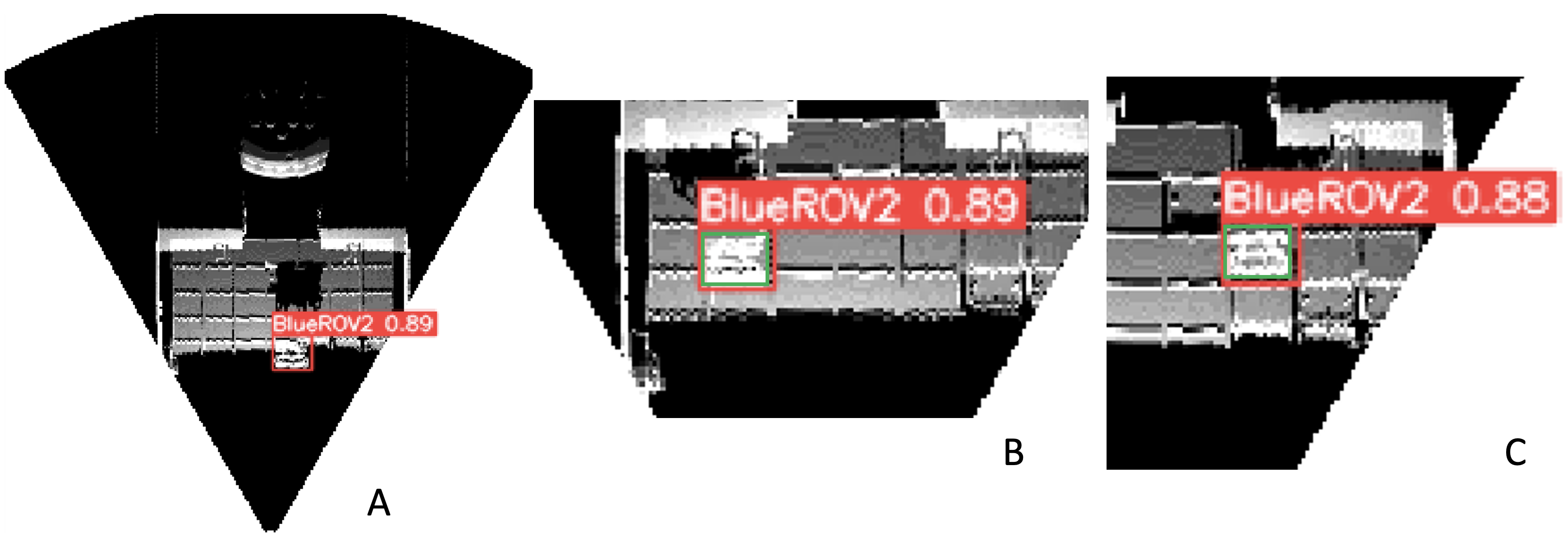}
    \caption{Sonar image view with YOLOv5 object detection. (A)~Accurate ROV detection by YOLOv5 (B) and (C)~Inaccurate ROV detection by YOLOv5.}
    \label{fig:yolofault}
\end{figure}

As indicated in Table~\ref{tab: errors_single_axis}, the MED remains below 200~mm for each trajectory type. In datasets 2 and 3, the values are relatively higher, approaching 200~mm. This is attributed to the BlueROV2 passing through specific locations during the bouncing and lawnmower trajectories, where, in the sonar images, the spatial overlap between the ROV and other objects in the underwater environment occurs. Under these circumstances, YOLOv5 fails to provide accurate pixel coordinates for the ROV, as shown in Fig~\ref{fig:yolofault}. In Fig~\ref{fig:yolofault}A, the ROV is not significantly affected by the environment, allowing YOLOv5 to clearly detect the BlueROV2 in the sonar images. However, in certain areas, as shown in Fig~\ref{fig:yolofault}B and C, this is not the case. The green bounding box indicates the expected range, while the red represents the results from YOLOv5.

\subsection{Experimental hardware architecture}
In the experimental validation of CAP-SD, the ASV was MallARD. A multi-beam sonar (Oculus M3000d) was mounted onto the chassis of the ASV, while self-localisation relied on the Sick TiM571 LiDAR and Microstrain 3DMGQ7 IMU. The communication link to the base station was made tetherless. The underwater ROV used in thsi work was the tethered BlueROV2.

\begin{figure*}[ht]
    \centering
    \includegraphics[width=\textwidth]{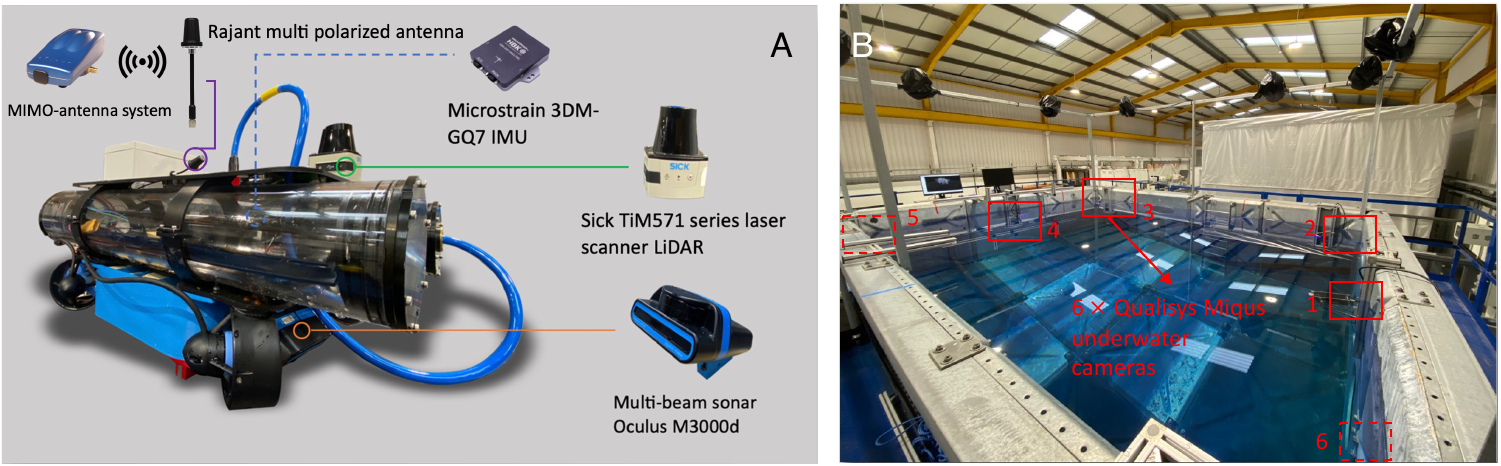}
    \caption{Hardware and System Architecture: Layout of MallARD components (left) and experimental field (right).}
    \label{fig:exp_hardware}
\end{figure*}

The CAP-SD system was assessed with data collected from a mock-up of a NFP (4.8~m $\times$ 3.6~m $\times$ 2.0~m), which was located at the RAICo1 lab in Whitehaven, UK. The positioning accuracy of the proposed technique was validated using the Qualisys motion tracking system, which consisted of six cameras, as depicted in Fig~\ref{fig:exp_hardware}.

\subsection{Online positioning performance}
\begin{figure}[ht]
    \centering
    \includegraphics[width=\columnwidth]{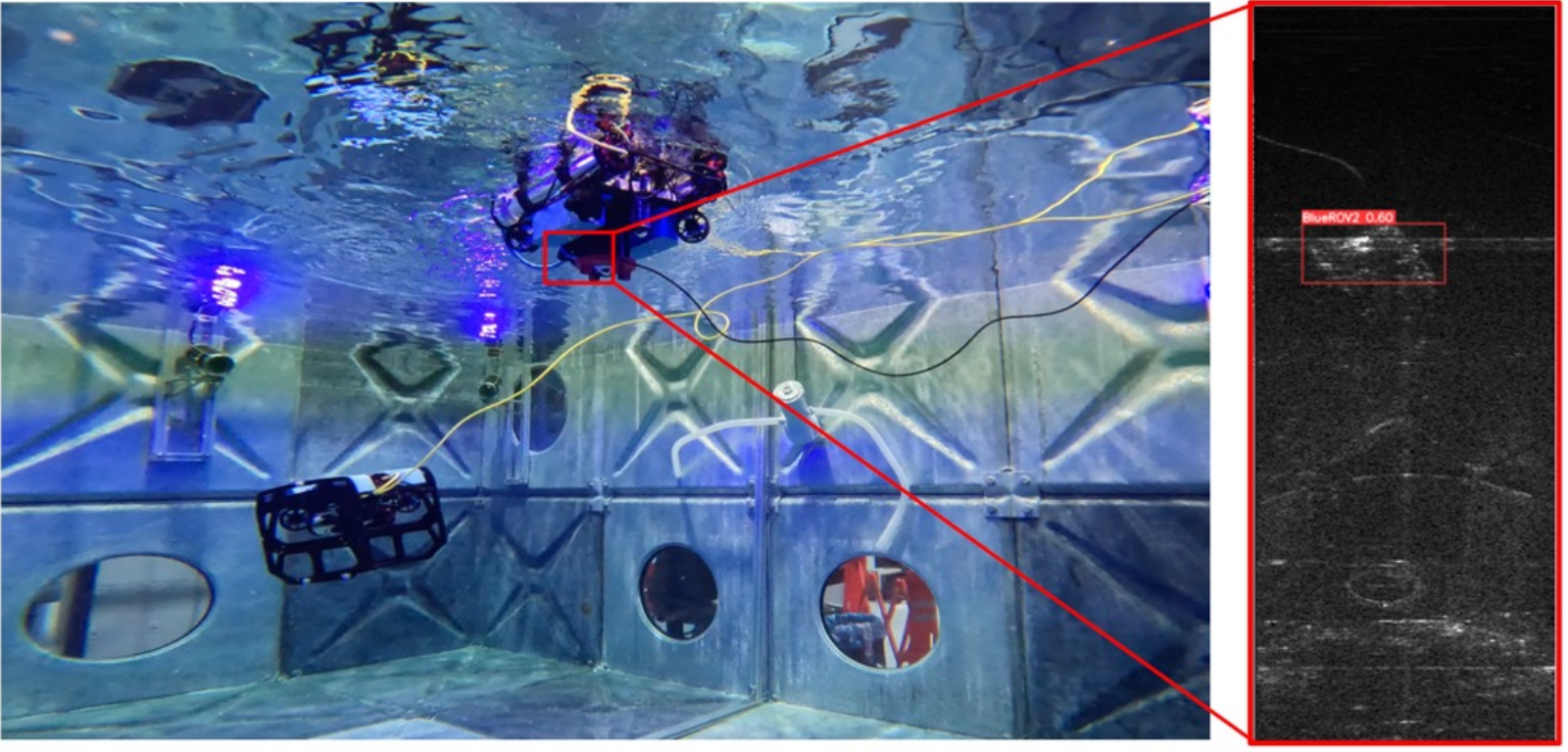}
    \caption{Underwater experiment setup of CAP-SD and the corresponding FOV of the multi-beam sonar.}
    \label{fig:capsd_yolo}
\end{figure}

To evaluate the performance of CAP-SD when using real hardware, the ASV and ROV moved synchronously within the mock-up of the NFP. Under manual control, the BlueROV2 performed translational and rotational motions in the X-Y plane, with depth changes within a 2-meter range. Three sets of data, each lasting approximately 80 seconds, were collected during validation experiments. The differences among the data sets were minimal, mainly in recording times, with all experiments conducted under manual control. To better integrate sonar images as one of the inputs for real-time positioning in the CAP-SD system, the official Oculus software~\cite{oculussoftware} was not used to display the sonar images (Fig~\ref{fig:nndetect}B). Instead, the corresponding ROS package was adopted. Additionally, YOLOv5m was applied in real time on the ROV through the ROS-YOLO interface. Fig~\ref{fig:capsd_yolo} presents a sample image of the BlueROV2 successfully detected by YOLO and an online visual representation of the ROV's position.

\subsection{Experimental results discussion}
\begin{figure}[ht]
    \centering
    \includegraphics[width=\columnwidth]{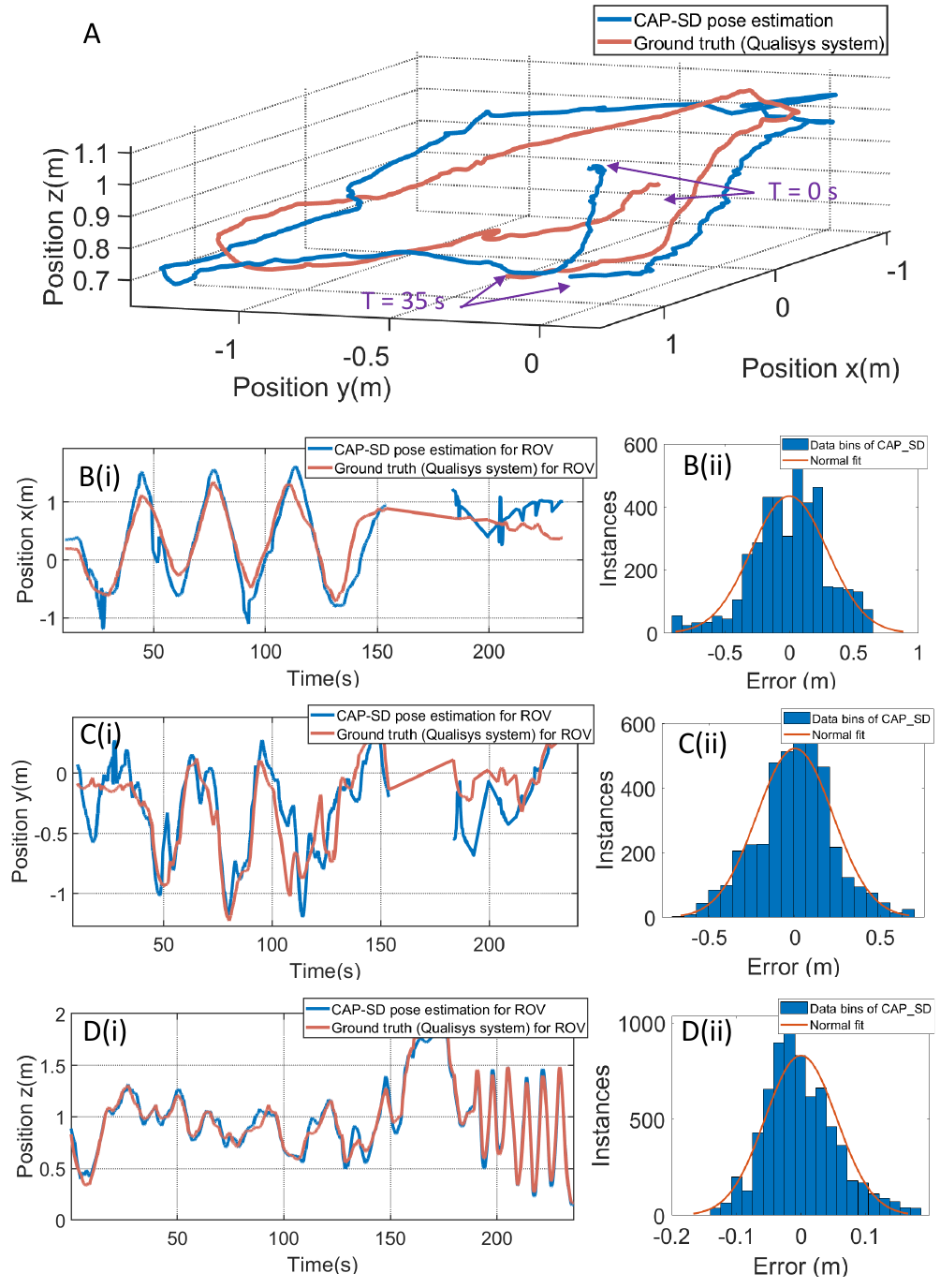}
    \caption{The positioning results of the CAP-SD system from the X, Y, and Z axes.\textbf{A(i)}, \textbf{B(i)} and \textbf{C(i)} The positioning of BlueROV2 by \mbox{CAP-SD} in comparison with the ground truth along the X, Y, and Z axes, respectively. \textbf{A(ii)}, \textbf{B(ii)} and \textbf{C(ii)} The error histograms of \mbox{CAP-SD} on the \textit{x}, \textit{y}, and \textit{z} axes.}
    \label{fig:capsd_xyz_exp}
\end{figure}

Fig~\ref{fig:capsd_xyz_exp}A displays the online positioning results of CAP-SD compared to the ground truth. Due to significant overlap in the ROV's trajectory over its entire dataset, it was not feasible to plot the full trajectory intuitively, so only 42 seconds of the trajectory are shown here (the full dataset can be found at~\cite{capsd}). Fig~\ref{fig:capsd_xyz_exp}B(i), C(i), and D(i) show BlueROV2's translation in the world frame, comparing CAP-SD and Qualisys (ground truth). 
Table~\ref{tab:capsd_exp_table} shows the full results for each dataset. The results indicate that the MED for CAP-SD ranged from 277.5~mm to 410.0~mm. Across all datasets, the MED was 331.1~mm. It was evident that the error along the $z$-axis was relatively small, which was due to the fact that depth information was primarily obtained from the depth sensor, which was highly accurate and the majority of the error resulted from estimation along the \textit{x} and \textit{y} axes.
One of the reasons for this error was that in the CAP-SD system, YOLO detected and locked onto the centre of the object. However, when the BlueROV2 experienced changes in depth or planar movement, its roll and pitch angles varied. Consequently, the depth sensor, mounted at the tail of the BlueROV2, did not accurately reflect the depth of the ROV's centre.

% \begin{figure*}[ht]
%     \centering
%     \includegraphics[width=\textwidth]{figures/abcdplots.png}
%     \caption{plots (A) and (C) shows the translation of the ROV in the the world fixed frame estimated using the two CAP-SD methods alongside the ground truth. Plots (B) and (D) show the associated error histograms for CAP-SD}
%     \label{fig:xyz}
% \end{figure*}

\begin{table}[ht]
    \centering
    \caption{Comparative performance of CAP-SD across various datasets.}
    \includegraphics[width=\columnwidth]{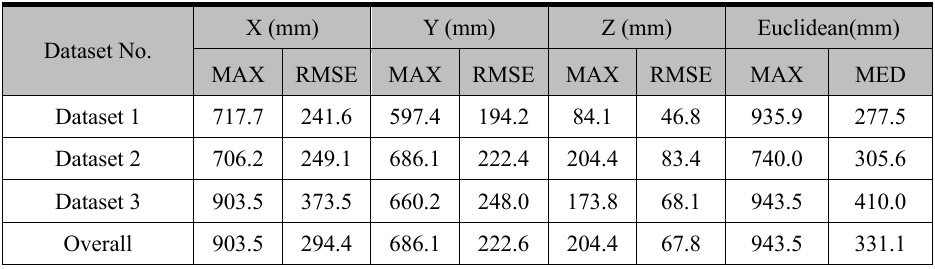}
    \label{tab:capsd_exp_table}
\end{table}

% \begin{table}[ht]
% \caption{RMS errors relative to the ground truth measurement of on CAP-SD for all 5 datasets. \label{tab: errors_single_axis}}
% \centering
% \begin{tabular}{ccccc}
% \toprule 
% \hline
% {\small{}Dataset~\tablefootnote{All datasets are available in:~\url{https://github.com/Xue1iang/IROS2024_CAP-SD}}}        & 

%%%%%%%%%%%%%%%%%%%%%%%%%%%%%%%%%%%%%%%%%%%%%%%%%%%%%%%%%%%%%%%%%%%%%

\section{Conclusions and future works}\label{sec:conclusion}
This paper introduced a novel underwater localisation system for joint robotic operations, capable of sim-to-real applications. To the authors' best knowledge, this is the first localisation system, applied in constrained underwater environments, that does not require infrastructure, is not limited by lighting conditions, and is unaffected by water turbidity. The paper validates the CAP-SD system through both simulation and experimental testing.

As well as demonstrating the potential of the CAP system, there are still several challenges that need to be overcome, including:
\begin{itemize}
  \item MallARD's autonomous tracking of the BlueROV based on object detection from the sonar image
  \item Fusing with a dead reckoning system to improve precision and account for outages in image sonar tracking
  \item Sensors' synchronisation in time domain
  \item A customized neural network designed specifically to detect objects in sonar images
\end{itemize}

This research offers significant potential for extension in several directions: 
\begin{itemize}
    % % \item Joint above-water and underwater 3D reconstruction in static/dynamic aquatic environments
    % \item Path planning for surface robots serving as external tracking systems to enhance the 3D SLAM capabilities of underwater robots
    % \item Temporal synchronisation of sensors distributed across multiple robotic agents

    \item Using neural networks improves the CAP-SD system's localisation capabilities, enabling it to determine the ROV's 6-DOF pose.
    \item Path planning for ASV serving as external tracking systems to enhance the underwater SLAM capabilities of ROV.
    
\end{itemize}

%%%%%%%%%%%%%%%%%%%%%%%%%%%%%%%%%%%%%%%%%%%%%%%%%%%%%%%%%%%%%%%%%%%%%

\section*{Acknowledgment}
This research was funded by EPSRC under grants: EP/P01366X/1, EP/W001128/1  and by an impact acceleration account secondment scheme which was jointly funded by the University of Manchester and EPSRC. Lennox acknowledges the support of the Royal Academy of Engineering (CiET1819/13). The authors would like to acknowledge the RAICo Fellowship Programme for providing support, resources, and valuable industry engagement opportunities, which significantly contributed to the development and application of this work.
%%%%%%%%%%%%%%%%%%%%%%%%%%%%%%%%%%%%%%%%%%%%%%%%%%%%%%%%%%%%%%%%%%%%%%%%%

\bibliographystyle{ieeetr}
\bibliography{07_reference_2023}

\end{document}